\documentclass[sigconf, nonacm]{acmart}

\usepackage{caption}
\usepackage{subcaption}

\graphicspath{{./Fig/}}

\begin{document}

\title{Knowledge Bases in Support of Large Language Models for Processing Web News}



\author{Yihe Zhang}
\affiliation{%
  \institution{University of Louisiana at Lafayette}
  \city{Lafayette}
  \state{Louisiana}
  \country{USA}
}

\author{Nabin Pakka}
\affiliation{%
  \institution{University of Louisiana at Lafayette}
  \city{Lafayette}
  \state{Louisiana}
  \country{USA}
  }

\author{Nian-Feng Tzeng}
\affiliation{%
  \institution{University of Louisiana at Lafayette}
  \city{Lafayette}
  \state{Louisiana}
  \country{USA}
}


\renewcommand{\shortauthors}{Zhang and Pakka, et al.}


\begin{abstract}
Large Language Models (LLMs) have received considerable interest in wide applications lately. 
During pre-training via massive datasets, such a model implicitly memorizes the factual knowledge of trained datasets in its hidden parameters. 
However, knowledge held implicitly in parameters often makes its use by downstream applications ineffective due to the lack of common-sense reasoning. In this article, we introduce a general framework that permits to build knowledge bases with an aid of LLMs, tailored for processing Web news. 
The framework applies a rule-based News Information Extractor (NewsIE) to news items for extracting their relational tuples, referred to as knowledge bases, which are then graph-convoluted with the implicit knowledge facts of news items obtained by LLMs, for their classification. 
It involves two lightweight components: 
1) NewsIE: for extracting the structural information of every news item, in the form of relational tuples; 
2) BERTGraph: for graph-convoluting the implicit knowledge facts with relational tuples extracted by NewsIE. 
We have evaluated our framework under different news-related datasets for news category classification, with promising experimental results.

\end{abstract}
\maketitle

\section{Introduction}

\begin{figure}
    \centering
    \includegraphics[width=0.45\textwidth]{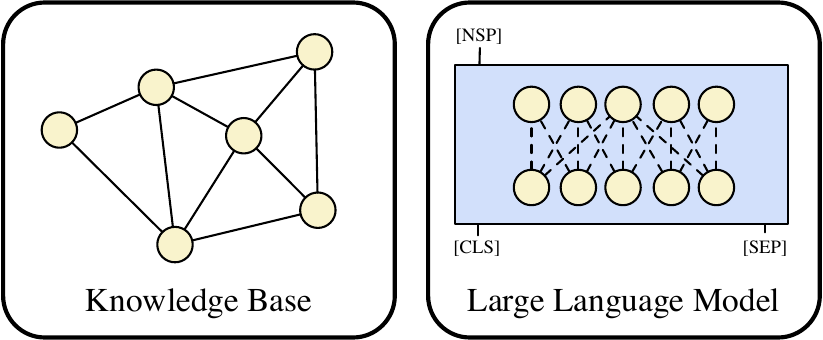}
    \caption{Illustration of Knowledge Base (left) and Large Language Model (right), 
    with the former typically storing structured knowledge explicitly and the latter holding unstructured knowledge implicitly.}
    \label{fig:KBLM}
\end{figure}

Large Language Models have garnered widespread enthusiasm lately as conversational agents for diverse applications. 
Noticeably, ChatGPT~\cite{A1:OpenAI} takes queries and creates responses, Google’s Bard~\cite{A2:Bard} can produce creative replies due to its coding and multilingual capabilities, ChatSonic~\cite{A3:Chatsonic} generates update-to-date replies with the aid of Google Search for accurate and informative content creation, among others. 
Meanwhile, a simple and yet powerful language representation model, known as Bidirectional Encoder Representations from Transformers (BERT)\cite{kenton:19:bert:NAACL}, was introduced. As an LLM family member, BERT comprises a stack of transformer encoders and is pre-trained via unlabeled texts. Most LLMs are fine-turnable using small amounts of domain-specific data for improving the performance of the given domain tasks.
In particular, the BERT model can be fine-tuned with just one additional output layer to create state-of-the-art models for a wide range of tasks~\cite{kenton:19:bert:NAACL}, such as question answering and language inference, with little changes to the models.

Conventional LLMs usually take massive amounts of unstructured text data for pre-training, before made them available for general applications. Although a pre-trained LLM can produce replies to its inputs (such as Web news items), it does not inherently capture structural and relational information of tokens existing in its every input, presenting an opportunity for improving the quality of replies. To this end, we resort to knowledge bases (KBs), which are created by pre-processing Web news items individually to extract structural and relational information of token in each item, such as the triggers, the arguments, the temporal relations, etc. This KB creation is undertaken in the rule-based manner automatically without requiring any relation-specific training data. Pre-processing news items extracts a number of relational tuples ($Arg1$, $Pred$, $Arg2$) per item, realized by the News Information Extractor (NewsIE) we have developed. The KBs created out of news items make it possible to produce better replies to inputs (i.e., news items) that are fed to a given LLM for processing, by complementing the LLM outputs. This results from the fact that LLM outputs contain no structural information while the KBs hold structural and relational information of individual news items.

In this article, we consider the framework that lets KBs support LLMs for improving Web news processing performance. The framework consists of an LLM and a graph convolutional network (GCN)~\cite{A9:defferrard:16:convolutional}, which is inputted with the LLM output and KBs for graph-convolutive operations to let KBs complement the LLM output. The GCN serve to convolute relational tuples with the implicit knowledge facts obtained by the LLM, which is fine-tuned by Web news items for improving the performance of news processing. The fine-tuned LLM takes raw Web news themselves directly as its input. While any LLM may be adopted to form the framework, we use BERT~\cite{kenton:19:bert:NAACL} as an example LLM in this article, realized our BERTGraph. Without making any change to BERT for use in BERTGraph, our framework is in contrast to prior designs, like KG-BERT~\cite{yao:19:kg:arXiv} and K-BERT~\cite{liu:20:kBert:AAAI}, which change the BERT input format, and Relphormer~\cite{bi:23:relphormer:arXiv}, which modifies the BERT encoder structure. 
However, any change to the input format or the encoder structure of BERT destroys its original embedding, requiring prohibitively expensive LLM model re-training, making our BERTGraph more favorable in practice and able to fully utilize implicit knowledge learned by BERT. More details of the BERTGraph framework can be found in Subsection~\ref{subsec:model} For the BERT output and KBs to be compatible with the GCN input, a text-to-graph adapter is devised, as stated in Subsection~\ref{subsec:t2g}.

We have implemented BERTGraph for experimentally evaluating its news category classification performance under three publicly available Web news datasets, N24News~\cite{wang2021n24news}, SNOPES~\cite{Snopes}, and politifact~\cite{Politifact}. Our evaluation results demonstrate that BERTGraph outperforms its BERT counterpart for evaluated news datasets, in terms of all performance metrics but the precision of politifact, with $0.27$ versus $0.28$.

The remainder of this paper is organized as follows. Section~\ref{sec:related} presents related background and prior work. Section~\ref{sec:bertgraph} overviews the proposed BERTGraph design, with its two key components detailed in sequence. Section~\ref{sec:exp} describes the implementation of BERTGraph and then provides experimental results under different news datasets for news category classification. Section~\ref{sec:conclusion} concludes this paper.

\section{Related Background and Prior Work}
\label{sec:related}
Background and prior work related to our proposed BertGraph design include knowledge bases, large language models, graph convolution networks, and information extraction, as provided next in sequence.

\subsection{Knowledge Bases}
Knowledge bases (KBs) are structured repositories of information and data. Such a repository contains facts and relationships among its entities. The Resource Description Format (RDF)~\cite{B12:mcbride:04:resource:HO} has been considered to be the standard format for knowledge bases in Natural Language Processing (NLP).
Various KBs have been created out of different data sources, and they are not all of high quality. In particular, DBpedia~\cite{B6:lehmann:15:dbpedia} created a multilingual knowledge base, extracted from Wikipedia. It uses the DBpedia ontology, comprising 320 classes, to store the facts in the RDF format. However, the lack of a consensus on the contributors of DBpedia poses an issue about its quality~\cite{B7:nguyen:16:type:ISWC}.
WordNet~\cite{B8:miller:95:wordnet} is an online database, which is an effective combination of traditional lexicographic information and modern computing. 
Parts of Speech of English are organized into sets of synonyms, with their semantic relations linking the synonym sets. 
Later knowledge bases, like YAGO~\cite{B4:fabian:07:yago}, are of higher quality due to the addition of knowledge about individuals like persons, organizations, products, etc., with their semantic relationships also in existence. 
YAGO claims its accuracy approaching 95\%, based on empirical evaluation of fact correctness. 
Meanwhile, DEAP-FAKED~\cite{B5:mayank:22:deap:ASONAM} and 
Boshko et al.~\cite{B10:koloski:22:knowledge:Neurocomputing} use an existing knowledge base, Wikidata5m, for Named Entity Disambiguation (NED). Embedding methods, such as TransE and ComplEX, are used to obtain embeddings for news classification.

The proposed BERTGraph framework relies on quality knowledge bases that are generated out of Web News datasets automatically, by our devised News Information Extractor (NewsIE). Generating knowledge bases at scale without manual effort, NewsIE is detailed in Section 3.2.

\subsection{Large Language Models}
With the introduction of Transformer~\cite{B11:vaswani:17:attention:neuIPS} in 2017, the NLP field has seen a rising number of efficient language models. 
Pre-trained Large Language Models (LLMs), like ChatGPT~\cite{A1:OpenAI}, Bard~\cite{A2:Bard}, ChatSonic~\cite{A3:Chatsonic}, BERT~\cite{kenton:19:bert:NAACL}, XLNet~\cite{B2:yang:19:xlnet:NeuIPS}, RoBERTa~\cite{B3:liu:19:roberta:arXiv}, etc., have been used extensively for NLP and other tasks. Since pre-trained models sll contain huge amounts of textual information, they can handle a variety of tasks easily and often effectively by down-streaming the models for tasks at hand, especially after model fine-tuning via task-specific datasets.

Bidirectional Encoder Representation from Transformers (BERT) \cite{kenton:19:bert:NAACL} has been fine-tuned for NLP tasks, such as sentiment analysis~\cite{B13:hoang:19:aspect:NCCL, B14:xu:19:bert:arXiv}, text classification~\cite{B10:koloski:22:knowledge:Neurocomputing, B15:sun:19:fine:CCL, B16:rai:22:fake}, etc. Upon training, BERT introduces masks to input sentences and predicts masked tokens. Due to its reliance on masks, BERT neglects dependency between the masked tokens positions~\cite{B2:yang:19:xlnet:NeuIPS}. XLNet~\cite{B2:yang:19:xlnet:NeuIPS} overcomes the limitation of BERT using its autoregressive formulation. BERT is trained modestly to achieve sound results~\cite{B3:liu:19:roberta:arXiv}. RoBERTa ehnances BERT by training with bigger batches of more data, dynamicaly changing the masking pattern, training on longer word sequences, and removing the next sentence prediction. Although pre-trained models are capable for NLP tasks, they are not always efficient for knowledge-driven tasks due to the discrepancy of fine-tuning’s specific domain and pre-training’s wide domains~\cite{liu:20:kBert:AAAI}.

\subsection{Graph Convolution Networks}
The generalization of convolutional neural networks (CNNs) to signals defined on more general domains was first attempted in~\cite{A8:bruna:13:spectral}, treating two onstructions of deep neural networks for processing graphic data efficiently. 
Later, gneralizing CNNs for use in high-dimensional irregular domains, such as graphs, was demonstrated in~\cite{A9:defferrard:16:convolutional} to learn local, stationary, and compositional features on graphs. 
Meanwhile, the Graph Convolutional Network (GCN) was proposed for semi-supervised learning on graph-structured data, based on a variant of CNNs~\cite{A10:kipf:16:semi:arXiv}. 
It scales linearly with the graph edge count and learns hidden layer representations that encode both the local graph structure and node features node, to yield superior performance. 
Relational Graph Convolutional Networks (R-GCNs) were introduced specifically to handle the highly multi-relational data characteristic of realistic knowledge bases, able to soundly outperform their decoder-only baselines. 
A comprehensive review on GCNs is made by grouping existing models into two categories, based on the types of convolutions, and also by categorizing them according to the areas of their applications~\cite{A11:zhang:19:graph}.

\subsection{Information Extraction}
Various information extraction (IE) mechanisms have been considered. Specifically, TextRunner~\cite{yates:07:textrunner:NAACL} pursues Open Information Extraction (OIE), that makes a single, data-driven pass over the entire corpus to extracts a large set of relational tuples autonomously. It utilizes a set of patterns in order to obtain propositions but does not capture the `context' of each clause for effective extraction. A follow-up study relies on semantic features (semantic roles) for the OIE task, demonstrating that Semantic role labeling (SRL) can be used to increase the precision and recall of OIE~\cite{christensen:10:SRLIE:NAACL}. Separately, a greedy parser, which relies on a classifier to predict the correct transition based on a small number of dense features, is treated for speedy parsing [6]. Subsequent OIE systems include Stanford OPENIE~\cite{A5:angeli:15:leveraging}, OPENIE4~\cite{A6:mausam:16:open}, and Neural OIE~\cite{A7:cui:18:neural:arXiv}, which have been used in various applications, alebit to their abilities on sentence level extraction only.

The OpenIE methods are in three lines.
The first lines of the methods start from TextRunner~\cite{yates:07:textrunner:NAACL},WOE~\cite{wu:10:WOE:ACL}, and OLLIE~\cite{schmitz:12:OLIE:EMNLP}, using traditional machine learning methods to learn the informative relation structures. For example, TextRunner extract unlexicalized Part-of-Speech (POS) and noun phrase (NP) chunk features and input them into the Naive Bayes classifier. WOE trains a linear-chain Conditional Random Field model by learning POS tags and Dependency Parsing (DP) tags extracted from the text.
OLLIE relies on bootstrap learning of patterns based on dependency parse paths. 
Different from the first line of simple machine learning methods, 
the second line uses deep learning methods for information extraction based on deep features.
LSTM-based models are first explored in this area by considering the information as a sequence, such as BiLSTM-CRF~\cite{huang:15:BiLSTM-CRF:arXiv},  BiLSTM-CNNs-CRF~\cite{ma:16:BiLSTM:arXiv} and Freedom~\cite{lin:20:freedom:KDD}.
Table Convolutional Networks is proposed~\cite{wang:21:tag:arXiv, wang:21:tcn:WWW} for extracting tablet information from the Web.
Recently, BERT-based models have emerged by formulating the IE problem as structural reading comprehension, such as WebSRC~\cite{chen:21:WebSRC:EMNLP},
and webformer~\cite{wang:22:webformer:WWW} build on transformer aims to extract HTML patterns.
However, these machine learning methods usually require a labeled dataset targeted to a specific domain.
The third line of OpenIE is rule-based methods.
This line of methods does not rely on any training data but only on previously defined rules.
Reverb~\cite{fader:11:Reverb:EMNLP}, extract manually selected syntactical and lexical features.
KRAKEN~\cite{akbik:12:kraken:AKBC} is designed for capturing complete facts.
The clause type is defined in ClausIE~\cite{del:13:ClausIE:WWW}
PROPS~\cite{stanovsky:16:props:arXiv} and PredPatt~\cite{white:16:PredPatt:EMNLP} parsing the information based on a list of rules defined on Universal Dependency.
Rule-based methods do not require any training process and can easily operate on large-scare datasets.

\section{BERTGrpah Framework and Knowledge Bases}
\label{sec:bertgraph}
\begin{figure*}[t]
    \centering
    \includegraphics[width=0.9\textwidth]{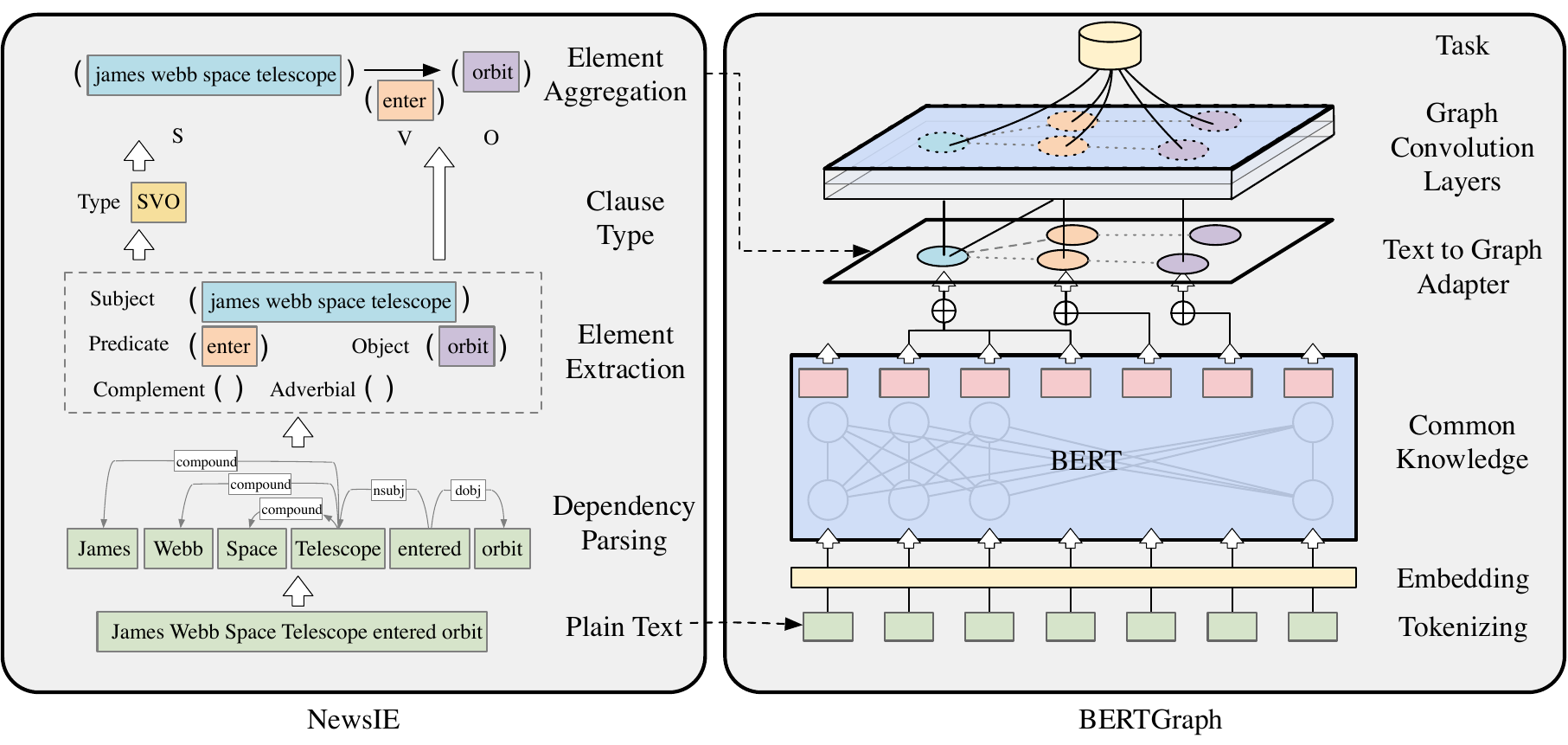}
    \caption{Overview of BERTGraph framework.}
    \label{fig:overview}
\end{figure*}

This section first overviews the proposed BertGraph framework, which comprises an LLM (e.g., BERT~\cite{kenton:19:bert:NAACL}) followed by a GCN, as shown in Fig.~\ref{fig:overview}. 
A key component in support of BERTGraph, called News information extractor (NewsIE), is then detailed. 
NewsIE is responsible for extracting structural information out of every Web news item individually to constitute the quality knowledge bases of news datasets. 
Such a knowledge base and the BERT's output are fed to the GCN for processing.

\subsection{BERTGraph Overview}
Pre-trainned LLMs show robust performance when fine-tuned with only some domain-specific examples [5] before putting them for use. 
Such robust performance nature stems from models’ implicit knowledge. However, materializing the implicit knowledge of a given LLM (e.g., BERT) is challenging. This work proposes to address the challenge by appending a Graph Convolution Layer to BERT, constituting BERTGraph. 
The Graph Convolution Layer is realized by a typical GCN, which is inputted with BERT’s output plus the dataset-specific knowledge base produced automatically by our NewsIE. Our BERTGraph framework aims to answer the key question of: Can structured KBs enhance the performance of LLM’s downstream tasks, given that LLM’s ouputs are unstructured? To this end, we provide the model and architecture of BERTGraph, in Sec~\ref{subsec:model}, followed by a light-weighted solution for dealing with the incompatibility issue of the BERT output format and the GCN input format, called the Text-to-Graph Adapter, in Sec.~\ref{subsec:t2g}.

\subsubsection{Model and Architecture}
\label{subsec:model}
BERTGraph takes the BERT Base [28] as its backbone. It should be noted that all other transformer-based LLM models are also suitable for BERTGraph, making it a general design framework. BERT consists of transformer encoders that focus on understanding the meanings of its inputted texts, sufficing our purpose of news category classification. The BERT Base is a pre-trained language model that leverages 12 transformer encoder layers to capture semantic-rich information from inputted texts. Its pre-training is on a large-scale unlabeled general domain corpus. Each input (with no more than 512 words) fed to BERT is first split into a list of tokens, by a tokenizer (which comes with a pre-trained BERT model), to become compatible with the BERT architecture. A token first transfers to an input embedding by adding its 1) token embedding, 2) segment embedding, and 3) positional embedding, where the token embedding captures the semantics of individual tokens, the segment embedding distinguish tokens across different inputs, and the positional embedding indicates the positions of the tokens situated in the input. An input embedding can then be forwarded to BERT for encoding, starting from its first transformer encoder layer. The stacked transformer layers of BERT learn the relationships among input tokens and output the following: 1) hidden states, which represent the contextualized representations and attentions, and 2) Attention scores, which denote attention probabilities assigned to input tokens. Notice that BERT outputs do not contain any structural information.

Without making any change to BERT, BERTGraph adds structure information, derived from NewsIE, directly to the BERT output, for graph-convolutive processing. Hence, the BERTGraph architecture is in contrast to prior designs, like KG-BERT [71] and K-BERT [38], which change the BERT input format, and Relphormer [3], which modifies the BERT encoder structure. The rationale behind our architecture is as follows. First, any change to the input format or the encoder structure destroys the original embedding, requiring prohibitively expensive LLM model re-training. Second, an input format chance calls for developing a new suitable data format, possibly a daunting task even for a specific domain. Conversely, BERTGraph makes full use of implicit knowledge learned by BERT, by forwarding its outputs, together with structure information (created by NewsIE; see Sec. 3.2 below), to the appended GCN for graph-convolutive processing. For the BERT output to be compatible with the GCN input, a text-to-graph adapter is devised, as stated next.

\subsubsection{Text-to-Graph Adapter}
\label{subsec:t2g}
\begin{figure}
    \centering
    \includegraphics[width=0.4\textwidth]{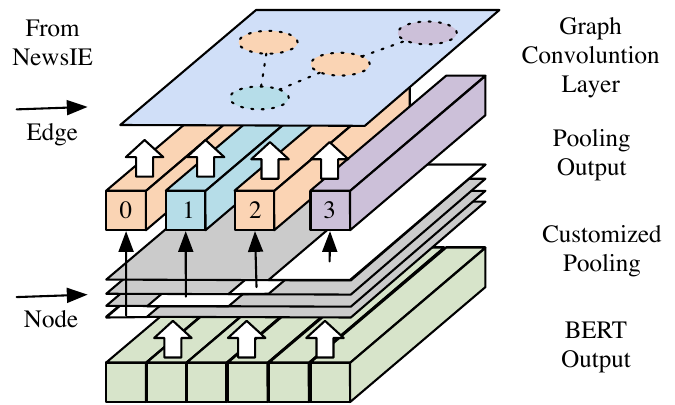}
    \caption{The design of Text-to-Graph Adapter.}
    \label{fig:t2g}
\end{figure}
The Text-to-Graph (T2G) adapter indispensable for BERTGraph is shown in Fig. 3. It takes the BERT’s output of a given input, plus the knowledge base produced out of the input (by NewsIE), as its input. The T2G Adapter generates Customized Pooling Layers, according to the knowledge base, by first examining those distinct nodes present therein. For each examined node, a dedicated pooling layer is generated by T2G Adapter, called the Customized Pooling Layer, which averages the token embeddings of its associated nodes, where the embeddings are from the BERT outputs, as depicted in Fig.~\ref{fig:t2g}. If there are a total of $K$ distinct nodes extracted by NewsIE, T2G Adapter creates the Customized Pooling Layer $K$ times. Each pooling output is then assigned with a new label by the T2G Adapter, starting form 0 and with an increment of $K$. This helps to train the appended GCN in the batch mode upon BERTGraph fine-tuning, when the GCN re-indexes knowledge bases. Finally, the pooling output and the edge list (present in knowledge bases) serve as the GCN's input.

\subsection{News Information Extractor}
\label{subsec:newsie}
As a key component that supports our BertGraph framework, the News Information Extractor (NewsIE) aims to identify the informative and structural aspects of each raw Web news item individually, to arrive at knowledge bases. NewsIE operates in a customized rule-based manner, relying on a set of rules derived from traditional NLP taggings, such as Dependency Parsing (DP)~\cite{de:21:UD:CL} and Part of Speech (POS)~\cite{petrov:12:UniPOS:LREC}. It automatically and efficiently identifies the informative and structural aspects of news items at scale, comprising predicate-argument structure extraction, clause type identification, and element aggregation. Specifically, the subject, predicate, object, complement and adverbial informative aspects of individual news items are identified and extracted. Next, based on the extracted informative aspects, NewsIE categories the extracted informative aspects per news item into their proper clause types. Finally, the structural aspects of each news item are obtained by applying our developed rules to its categoried clause types. Realized by using Python NLP library SpaCy~\cite{honnibal:20:spacy}, our NewsIE is detailed below.

\subsubsection{Predicate-argument Structure Extraction}
\label{subsec:extract}
Web news evolve constantly, with new subjects and events emerging daily. However, there are two drawbacks if deploying LLMs directly for processing Web news. 
First, LLMs require substantial computational resources in their training, making it impractical to retrain them even after abundant new subjects and events have emerged. Second, LLMs lack scalability when exploring previously unknown subjects or events. For instance, the BERT model assigns an `UNKNOWN' token to any word non-existing in its dictionary (upon prefix search failures). 
To tackle these drawbacks, we extract factual information from the unstructured text of each news item as pre-processing, bapplying Part-of-speech tagging (POS) to get token-level grammatical category information, e.g., `NOUN' and `VERB'. This factual information extraction is inspired by the Open Information Extraction (OpenIE) technique. Following the rule-based OpenIE method, clausIE~\cite{del:13:ClausIE:WWW, chourdakis:18:ClausIE:DMRN}, our structure extractor employs dependency paring (DP)~\cite{honnibal:15:EMNLP:DEP} to get token-level relationship tagging. Note that Stanford DP (StanDP)~\cite{chen:14:StanfordDP:EMNLP} may be suitable for extraction as well. Unless stated otherwise, we adopt DP as the default method. After DP, a plain news sentence is transferred to a tree-like syntactic structure, where each node is a word in the sentence, and edges or arcs between pairs of nodes represent grammatical relationships. The root node of the whole tree structure is usually the main verb of the sentence. However, if the verb doesn’t exist, the first noun (or the subject complement of a copular verb in StanDP) of the sentence becomes the root node. A group of nodes within a sub-tree, which denotes one single concept or synonym, is called a chunk. The goal of our NewsIE is to extract the subject, predicate, object, complement, and adverbial chunks from news items.

Given that subjects are indispensable in most news items, we start by extracting subjects. From the root node, we look for the node in the dependency tree labeled as either `NSUBJ' (subject), `CSUBJ' (clausal subject), `NSUBJPASS' (passive subject), and `CSUBJPASS' (clausal passive subject), as the subject head. 
Notice that multiple subject heads extracted from the single dependency tree are possible. 
If no `VERB' exists in a sentence, the `ROOT' node serves as the subject head. We search the sub-tree for each subject head by traversing the compound (tagged as `COMPOUND') relationship, or `NOUN' (person, place, thing, animal, or idea) and `PROPN' (name or place) POS tag, recursively in its child nodes. Words tagged with `NOUN' and `PROPN' are open-class words that can be connected to make larger chunks. During travasal, eligible child nodes must be connected to the subject head (one or more hops) only with a compound relationship or belonging to any of the ‘NOUN’ and `PROPN' POS tags. This rule separates the hierarchy subjects that may exist in the same sub-tree. After obtaining the search results, we sort the nodes according to their original order in the text and consider them together as a subject chunk. Since DP and stanDP may produce ambiguous parsing results~\cite{krasner:22:reDP:arXiv}, we apply Named Entity Recognizer (NER), supported by SpaCy Stanza~\cite{qi:20:stanza:AMACLSD}, to rectify the results.

Then, we extract predicates from the sentences of each news item. Predicates are usually verbs (words with `VERB' POS tag) or copulas (words with `AUX' POS tag) in a sentence. From the dependency tree, we can easily locate the predicate for each subject chunk by checking the parent node of the subject head, denoted as the predicate head. Besides the single-word verb, six different cases of a predicate chunk are considered: 
1) Phrasal verb, formed with a verb and an adverb particle, e.g., `take off'; 
2) Prepositional verb, formed with a verb and a preposition, e.g., `worry about'; 
3) Phrasal-prepositional verb, formed with a verb, a particle, and a preposition, e.g., `catch up with'; 
4) negative verb, formed by a word with the `not' meaning and a verb, e.g., `never go'; 
5) tense verb, formed with ‘have’ or `be' and a verb, e.g., `is playing'; 
6) verb with passive voice, formed with `be' and a verb, e.g., `is done'. For Cases 1) and 3), NewsIE traverses the sub-tree of the predicate head, probing the node with a particle relationship that is tagged as ‘PRT’ in its child(ren). For Case 4), NewsIE simply finds the child node that has a negative relationship tagged as `NEG' with the predicate head. For Cases 5) and 6), NewsIE traverses the sub-tree to find the node respectively with auxiliary (tagged as ‘AUX’) and with passive auxiliary (tagged as `AUXPASS') relationships.

Notice that the valid search result of every case must be directly connected to its predicate head. All valid results (i.e., nodes) of a predicate head sub-tree are denoted as its predicate chunk. Since the prepositional verbs for Case 2) all have their semantic meanings very similar to that of their key verb, e.g., `worry' and `worry about', they are considered to be a single verb word when extracting the predicate head, with its prepositional words (e.g., `about' of `worry about') regarded as its objects.

As objects are associated with their respective verb chunks, NewsIE can extract objects from verb chunks. 
In general, the verb of a sentence may be followed by 
1) no objects, 
2) one object (direct object), 
3) two objects (direct object and indirect object), 
and 4) prepositional objects. To extract objects from a sentence, we start from the predicate head to extract its direct, indirect, and prepositional objects separately. 
In particular, extracting the direct (or indirect) object is to follow the node with a 1DOBJ' (or `DATIVE') DP tag but without an `ADP' (adposition) POS tag. 
The extracted direct (or indirect) object head then allows to get the subject chunks by searching nodes with ‘COMPOUND’ DP tags. A prepositional object is obtained by probing the sub-tree along a node with its DP tag being `POBJ' (preposition object), `PREP' (preposition), `AGENT' (agent), or `DATIVE' with `ADP', to serve as its POS tag. Notice that a sentence may have multiple prepositional object heads. 
The sub-tree of the obtained prepositional object head is then traversed along the node(s) tagged with ‘COMPOUND’ DP, to serve as the corresponding prepositional object(s). If no such tagged node is found, the verb has no linked object.

A complement refers to a predicate argument for completing the meaning of a predicate. 
Extracting the complement chunks starts from the predicate heads. 
For each head, we consider three types of complements, including `ATTR' for attribute, `OPRD' for object predicate, and `COMP' family for complements. The `ATTR' is a noun phrase usually following a copula verb.
The `OPRD' is an object predicate in a small clause that functions like the predicate of an object. 
The `COMP' family contains the complement of a preposition (tagged as `PCOMP'), the clausal complement (tagged as `CCOMP'), and the open clausal complement (tagged as `XCOMP'). 
The complement chunks of a predicate are derived by starting from the predicate head node to traverse the sub-tree along its child nodes with the aforementioned DP tags.

The last kind of elements to be extracted are adverbials, which contain such important information as space and time. Although many adverbials possibly exist in a sentence, we consider only those adverbials related to the predicate. 
In a dependency tree, adverbials usually start from the nodes tagged with DPs which include the substring of `ADV', i.e., adverbial clause modifiers (`ADVCL'), adverbial modifiers (`ADVMOD'), and noun phrases as adverbial modifiers (`NPADVMOD'). 
Hence, the complement chunks of a given adverbial can be obtained by probing its dependency tree’s nodes with the aforementioned DP tags.

\subsubsection{Clause Type Identification}
\label{subsec:clause}

\begin{figure}
    \centering
    \includegraphics[width=0.48\textwidth]{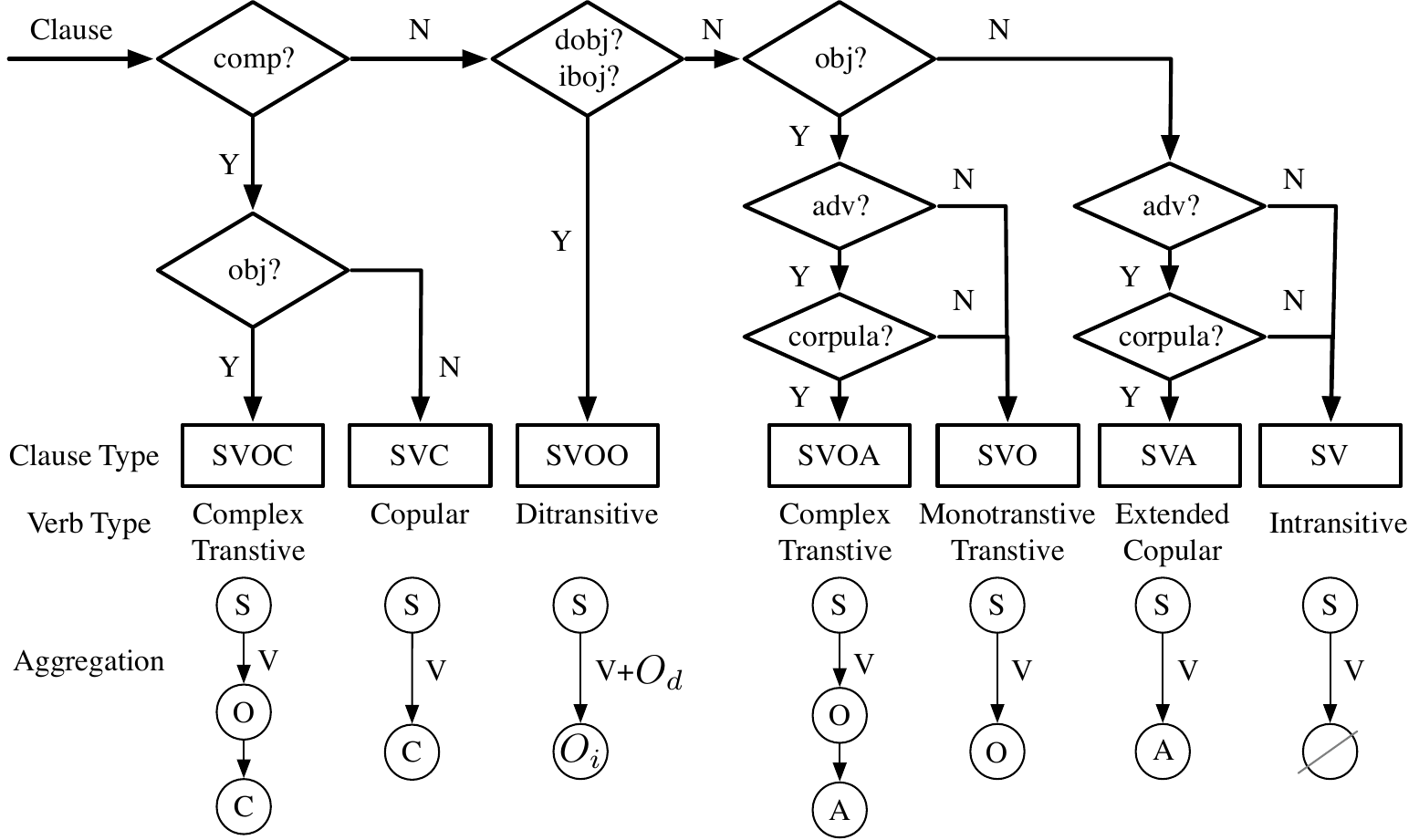}
    \caption{The sequence of clause type identification.}
    \label{fig:corpula}
\end{figure}

After extracting five kinds of chunks, i.e., subjects (`$S$'), predicates (`$V$'), objects (`$O_d$' for direct object, `$O_i$' for indirect object, `$O_p$' for prepositional object), complements (`$C$'), and adverbials (`$A$'), we next classify the types of clauses via a rule-based manner, where a clause comprises multiple extracted chunks. 
The clause type indicates how those extracted chunks are syntactically connected among one another. Before clause type identification for a given news item, we first apply VerbNet~\cite{schuler:05:verbnet} and PropBank~\cite{kingsbury:02:treebank:LREC} on the news item to categorize its verbs and get the argument frames of categorized verbs. We then map the extracted chunks of the news item to the argument frames given by VerbNet and PropBank. 
This way can quickly identify different types of verbs without managing an extensive dictionary of every kind of verbs. Given all the clause types contain `$S$' and `$V$', with their differences lying in the parts of `$O$', `$C$', and `$V$', we identify different clause types in the sequence shown in Fig.~\ref{fig:corpula}. 
First, if there is `$C$', we look for a ‘$O_d$’. 
If it exists but no `$O_i$' is present, the clause type of `$SVOC$' is designated. Otherwise, the clause type is of `$SVC$'. 
If both `$O_d$' and `$O_i$' are present, the clause type belongs to `$SVOO$'. 
In the case of `$O_d$' existing together with `$O_p$' or `$V$', the clause type is identified as `$SVOA$' if its verb is labeled as a linking verb according to VerbNet; otherwise, the clause type is of `$SVO$'. 
In the absence of both `$O_d$' and `$O_i$', the clause type belongs to `$SVA$', only if either ‘`$O_p$' or `$V$' exists and the verb is a linking verb. The remaining cases are assigned with the clause type of `$SV$'.

\subsubsection{Element Aggregation}
\label{subsec:aggregation}

After the clause types are identified, we then aggregate the chunks with different clause types, aiming to transform structured information into knowledge bases. 
Each clause type can generate one (or two) tuple(s). 
An element is denoted by a single node in the knowledge base. 
Since an `$SV$' clause involves only two elements, a dummy node (dented as `DUMMY'), which contains no information, is added to it, enabling its predicate node (`$V$') to link with its subject node ($S$) and the dummy node (`DUMMY'). 
For the clause type of `$SVO$', its predicate node (`$V$') is directly linked with its subject node (`$S$') and its object node (`$O$'). 
Likewise, for the clause type of $SVC$, its predicate node (`$V$') is directly linked with its subject node (`$S$') and its complement node (`$C$'). 
The clause type of `$SVA$' has its predicate node (`$V$') linked directly with its subject node ($S$) and its adverbial node (`$V$'). 
The `$SVOC$' clause type has its predicate node (`$V$') linked with its subject node ($S$) and its object node (`$O$'), with its complement node (`$C$') appended to the object node (`$O$'). 
Similarly, an $SVOA$ clause type has its predicate node ($V$') linked with its subject node ($S$) and its object node (`$O$'), with its adverbial node (`$A$') appended to object node (`$O$').
An `$SVOO$' clause type combines its predicate node (`$V$') and its direct object node (`$O_d$') to become a new node that denotes the relationship between the subject node ($S$) and the indirect object node (`$O_i$'). Hence, knowledge bases are generated automatically at scale without labeling effort.

\section{Experimental Evaluation and Result Discussion}
\label{sec:exp}
This section first introduces the datasets used in our evaluation study, followed by the experimental setup for experimental evaluation. 
Experimental results and discussion are then provided.

\begin{table*}
  \centering
  \caption{Comparative Results of BERTGraph and BERT on different datasets, under 10\% of data for training and 10\% for validation}
  \begin{tabular}{c|c|c|c|c|c|c|c|c|c|c|c|c}
    \toprule
    \multicolumn{1}{c|}{Dataset} & \multicolumn{3}{c|}{BERTGraph} & \multicolumn{3}{c|}{BERT} & \multicolumn{3}{c|}{BERTGraph-l} & \multicolumn{3}{c}{BERTGraph-s}\\
    \midrule
    Metrics & F1  & Acc & Pre & F1 & Acc & Pre & F1 & Acc & Pre & F1 & Acc & Pre\\
    \hline
    Snopes  & \textbf{0.68} & \textbf{0.71} & \textbf{0.68} & 0.66 & 0.66 & 0.67 & 0.67 & 0.7 & 0.66 & 0.6 & 0.66 & 0.58 \\
    N24News  & \textbf{0.68} & \textbf{0.69} & \textbf{0.67} & 0.64 & 0.64 & 0.65 & 0.65 & 0.65 & 0.66 & 0.56 & 0.57 & 0.58 \\
    Politifact  & \textbf{0.29} & \textbf{0.3} & 0.27 & 0.27 & 0.28 & \textbf{0.28} & 0.23 & 0.27 & 0.25 & 0.24 & 0.28 & 0.26 \\
    \bottomrule
  \end{tabular}
  \label{tab:tab1}
\end{table*}

\subsection{Datasets}
Three datasets N24News~\cite{wang2021n24news}, SNOPES~\cite{Snopes}, and politifact~\cite{Politifact} are considered for the evaluation of the proposed model. N24News is a multimodal news dataset containing both text and images from New York Times with 24 categories. The dataset is comprised of 60K image-text pairs, but this paper uses only the text part of the dataset. SNOPES is a rumor dataset consisting of a number of variety of news categories. Among the categories, only 12 categories are selected for the classification task totalling 7060 instances of news. Politifact is a fact-check dataset collected from PolitiFact website. There are 21152 instances and 6 categories of  expert verified facts.

\subsection{Metric}
We use F1-score (F1), accuracy (Acc), and precision (Pre) as evaluation metrics. Precision measures the accuracy of positive inference made by the model. In classification, it is the ratio of True Positive (TP) of a class to the sum of TP and False Positive (FP) of the same class. Recall is another metric which measures the ability of a model to correctly predict the positive instances. It is the ratio of TP to the sum of TP and False Negative (FN). F1-score is used to evaluate model on  imbalanced data. It is the harmonic mean of Precision and Recall which provides balance between them. Accuracy measures the number of correctly inferred instances from the whole dataset. 

\subsection{Experimental Setup}
The models were trained on DELL PRECISION TOWER 7910 containing 2 NVIDIA TITAN RTX TU102 GPUs with 24GB of memory each. Batch size of 16 is used with the learning rate of 1e-5 and Adam~\cite{kingma2014adam} is used as optimizer. 
The size of the training dataset varies, ranging from $5\%$ to $80\%$, with $10\%$ for validation dataset, and the remainder for test datset.
There are 4 layers of GNN with hidden layer dimension of 768. For evaluation, Accuracy is the default metric for classification tasks but due to its lack for imbalanced dataset~\cite{mayank2022deap}, F1 score is also considered.

\subsection{Experimental Results and Discussion}
We conduct a series of experiments on the three dataset, i.e., Snopes, N24News, and Politifact to evaluate the performance of our proposed BERTGraph.

\subsubsection{Comparing to the BERT baseline}
\label{exp:baseline}

To compare BERTGraph with BERT baseline, we employ various fine-tuning configurations.
Besides BERT and BERTGraph, we explore two variations of the BERTGraph model:
1) BERTGraph-s, this variant maintains the stability of BERT without conducting any fine-tuning;
2) BERTGraph-l, this variant, fine-tuning is applied solely to the last layer of BERT.
We first use $10\%$ of the data to fine-tune BERT, BERTGraph, BERTGraph-0 and BERTGraph-1, the results are shown in Table.~\ref{tab:tab1}.
From the table, it is evident that BERTGraph outperforms the other models on Snopes and N24News dataset across all performance metric (i.e., F1, Acc, and Pre).
Specifically, on Snopes dataset, BERTGraph achieves  F1, Acc, and Pre scores of $0.68$, $0.71$ of $0.68$, respectively,
whereas the counterpart BERT, only achieves $0.66$, $0.66$, and $0.67$ of F1 score, Acc, and Pre.
BERTGraph improvements of $2\%$, $5\%$, and $1\%$ in F1 score, Acc, and Pre, respectively.
On N24News dataset, BERTGraph achieves F1, Acc, and Pre scores of $0.68$, $0.71$ of $0.68$, 
making improvements of $4\%$, $5\%$, and $2\%$, respectively, over BERT model.
On the Politifact data, both BERTGraph and BERT exhibit performance below $0.3$ in terms of F1, Acc, and Pre.
However, BERTGraph outperforms BERT in terms of F1 and Acc, two metric, achieving values of $0.29$, and $0.3$, respectively.
The Pre score is slightly in inferior than BERT, at $0.27$, compared to $0.28$ for BERT.
The above results indicate that BERTGraph, outperforms BERT, since structural information helps BERT get better representations.
Now let's examine two variants of BERTGraph, i.e.,BERTGraph-l, and BERTGraph-s.
We can see BERTGraph-s performs the worst among all four models, the reason is that this model does not learns any structural information, which is crucial for improving performance.
BERTGraph-l performs slightly superior than BERT on Snopes and N24News dataset, but inferior than BERT on Politifact dataset.
These results suggest that fine-tuning only one layer in BERT can lead to improved performance in some cases.

Next, we conducted fine-tuning both BERTGraph and BERT by using $80\%$ of the dataset, the results are presented in Table.~\ref{tab:tab2}.
From the table, we can see BERTGraph consistently outperforms BERT across all F1, Acc and Pre metrics, on the Snopes, N24News, and Politifact datasets.
Specificlly, on the Snopes dataset, BERTGraph achieves F1, Acc, and Pre scores of $0.74$, $0.75$, and $0.74$, respectively,
whereas BERT only attains scores of  $0.68$, $0.69$, and $0.67$.
This demonstrates that BERTGraph surpasses BERT, up to $6\%$, $6\%$, and $7\%$, for F1, Acc and Pre, respectively, under $80\%$ of the training dataset setting.
For N24News dataset, BERTGraph maintains its superiority, with $6\%$, $6\%$, and $7\%$ higher than BERT, on the three metrics.
On Politifact, BERTGraph achieves the F1, Acc, and Pre, as $0.29$, $0.3$, and $0.3$, 
which outperforms $BERT$ by $6\%$,  $1\%$, and $2\%$, respectively.

\begin{table}
  \centering
  \caption{Comparative Results of BERTGraph and BERT on different datasets, under 80\% of data for training and 10\% for validation}
  \begin{tabular}{c|c|c|c|c|c|c}
    \toprule
    \multicolumn{1}{c}{Dataset} & \multicolumn{3}{c|}{BERTGraph} & \multicolumn{3}{c}{BERT} \\
    \midrule
    Metrics & F1  & Acc & Pre & F1 & Acc & Pre  \\
    \hline
    Snopes  & \textbf{0.74} & \textbf{0.75} & \textbf{0.74} & 0.68 & 0.69 & 0.67\\
    N24News  & \textbf{0.75} & \textbf{0.75} & \textbf{0.74}  & 0.69 & 0.69 & 0.70 \\
    Politifact  & \textbf{0.29} & \textbf{0.29} & \textbf{0.3} & 0.24 & 0.28 & 0.26\\
    \bottomrule
  \end{tabular}
  \label{tab:tab2}
\end{table} 

\subsubsection{Trianing Size impacts}
\label{exp:training}
\begin{figure}
    \begin{subfigure}[b]{0.43\textwidth}
        \centering
        \includegraphics[width=0.9\textwidth]{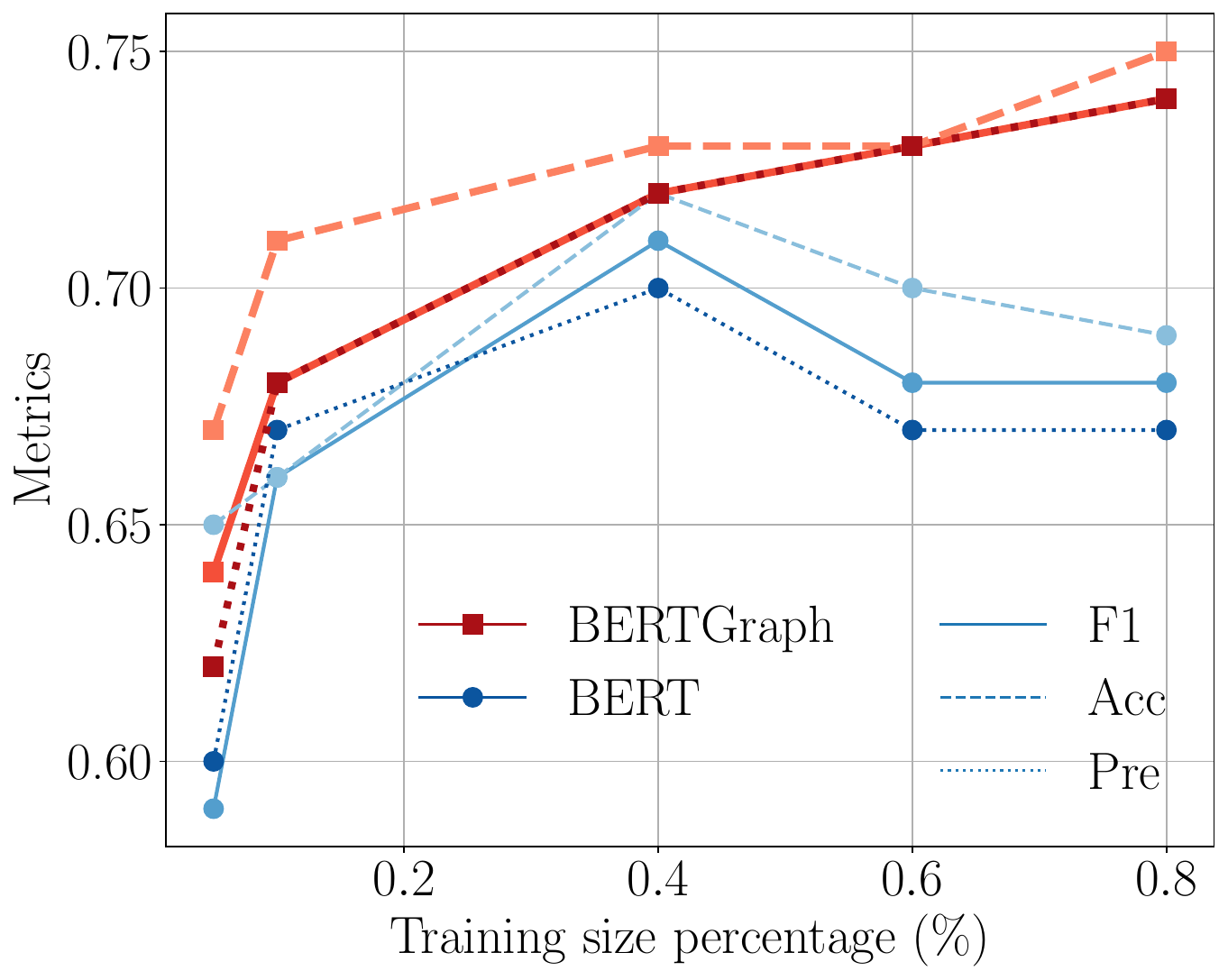}
        \caption{Snopes}
    \end{subfigure}
    \begin{subfigure}[b]{0.43\textwidth}
        \centering
        \includegraphics[width=0.9\textwidth]{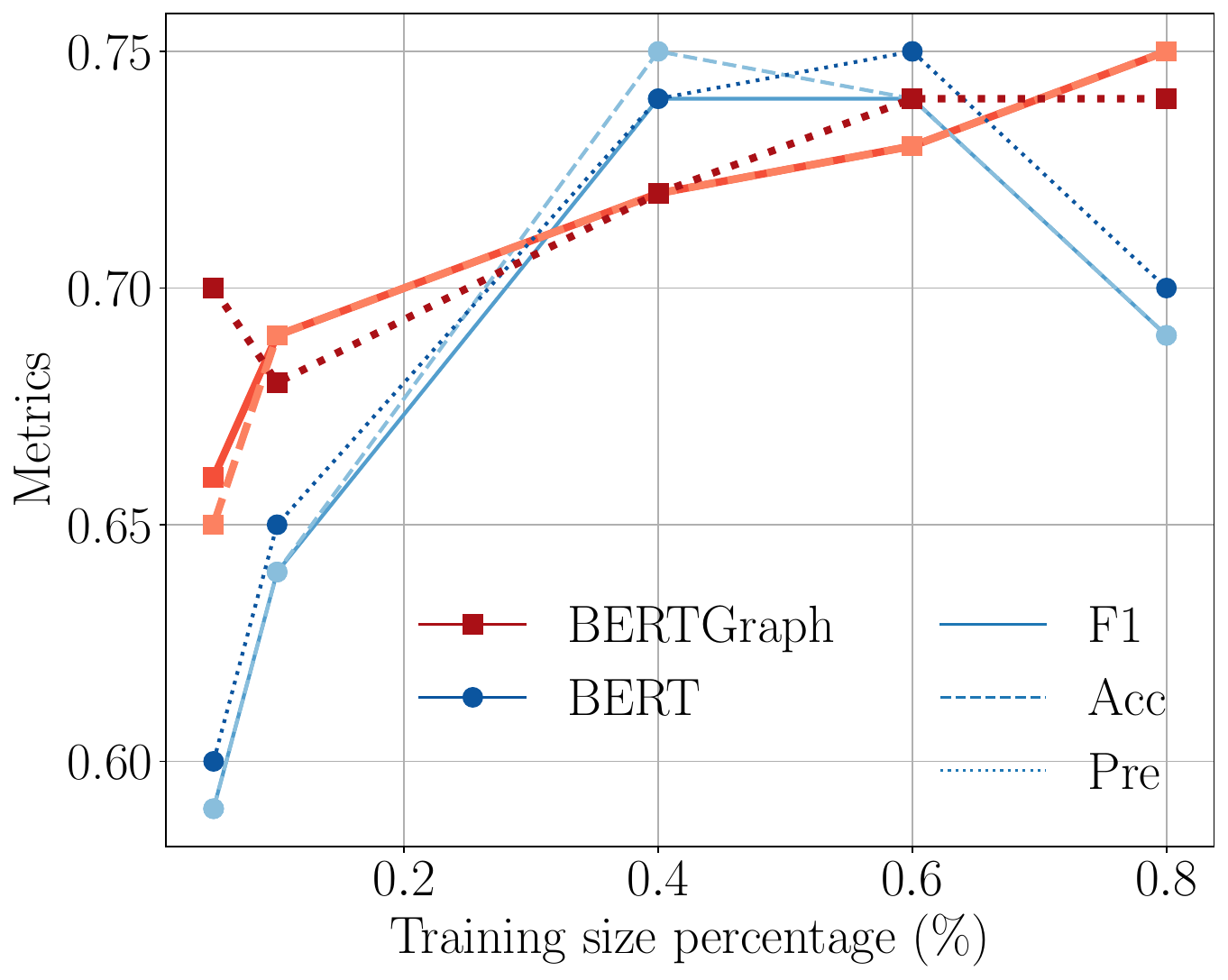}
        \caption{N24News}
    \end{subfigure}
    \begin{subfigure}[b]{0.43\textwidth}
        \centering
        \includegraphics[width=0.9\textwidth]{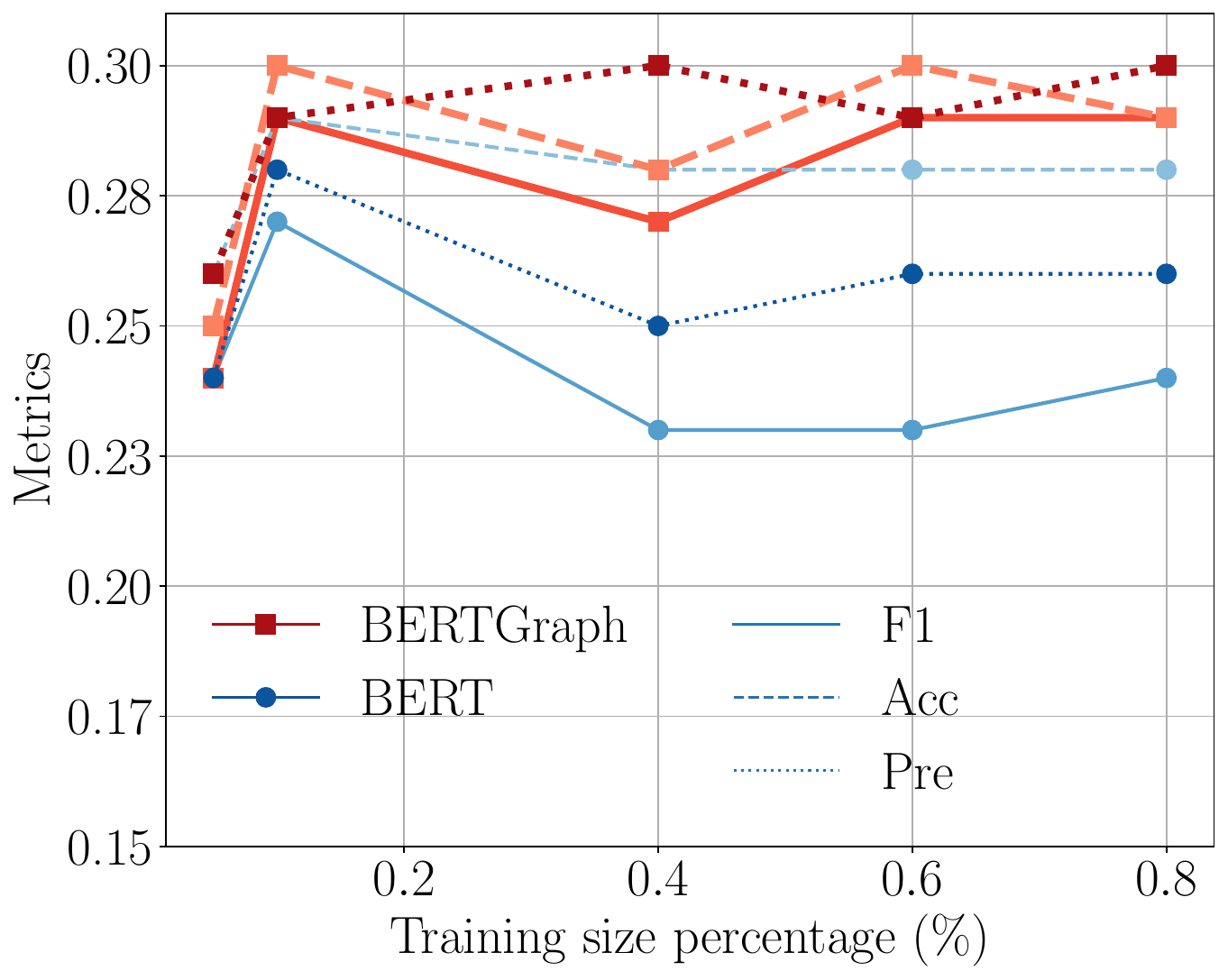}
        \caption{Politifact}
    \end{subfigure}
      
  \caption{F1 score, accuracy, precision under various training amounts (in \%) of datasets.}
  \label{fig:multi}
\end{figure}
A comparative study of the baseline model and BERTGraph on variable training size of datasets is shown in  Fig.~\ref{fig:multi}. It is clear that our proposed model performs better than the baseline on both F1 score and accuracy. But the difference is reduced, or even BERT performs better when the test and train datasets are nearly equal in proportion. However, this is rarely the case in real-world scenarios. The real-world problems have very little data to learn from. Sometimes, the model has to learn in one shot. The BERTGraph learns even from less amount of data due to the additional information obtained from NewsIE. The semantic and structural information obtained from the NewsIE focuses the model on attending to the important token rather than dividing the attention to all tokens. 
Fig.~\ref{fig:multi}(a) illustrates the BERTGraph and BERT performance on the Snopes dataset, where the training size varies from $5\%$ to $80\%$
The utilization of only $5\%$ of training data mimics the real-world scenarios where labeled data is scarce,
while the use of $80\%$ training data allows us to explore the overall learning capabilities of the models.
In Fig.~\ref{fig:multi}(a), the red lines with square markers denote the results from the BERTGraph model, 
and the blue lines with circular markers denote results from the BERT model.
The solid, dashed, and dotted lines correspond to F1, Acc, and Pre score, respectively.
Notably, BERTGraph outperforms BERT in nearly all the cases of the F1, Acc, and Pre scores.
BERTGraph shows an upward trend in performance as the training data size increases.
Whereas, BERT shows the performance improved initially but subsequently suffers a performance decline,
with scores for F1, Acc, and Pre decreasing from $0.71$, $0.72$, and $0.7$ to $0.68$, $0.69$, and $0.67$, respectively.
The reason behind this is fine-tuning BERT lets the model overfit the specific tasks.
However, BERTGraph do not suffer such problems according to the observation.
Fig.~\ref{fig:multi}(b) illustrates the BERTGraph and BERT performance on the N24News dataset.
The legends remain the same as the previous figure.
We observe that only a small portion of the data is available for fine-tuning.
When training data increase from $10\%$ to $40\%$ or $60\%$, the BERT model suddenly achieves its best performance.
Nevertheless, as the training data increased, the BERT model performance began to drop.
Specifically, its F1 drops from $0.74$ to $0.69$ , Acc decreases from $0.75$ to $0.69$, and Pre decreases from $0.74$ to $0.7$.
In contrast, BERTGraph stays an upward trend in performance, increasing its F1 score from $0.66$ to $0.75$, Acc from $0.65$ to $0.75$, and Pre from $0.7$ to $0.74$, when the training dataset increase from $5\%$ to $80\%$
This suggests fine-tuning with structural information can help mitigate the overfitting problem since noise may be ignored in the process.
Fig.~\ref{fig:multi}(c) illustrates the performance on the Politifact dataset.
The BERTGraph consistently outperforms the BERT model.
As the training size increase, BERTGraph shows slight improvements, with F1 increases from $0.24$ to $0.29$, Acc from $0.25$ to $0.29$, and Pre from $0.26$ to $0.3$

\begin{table}
  \centering
  \caption{Comparative accuracy of BERT based models and GCN based models on different datasets}
  \scalebox{0.8}{
      \begin{tabular}{c|c|c|c|c|c}
        \toprule
        Dataset & BERTGraph & BERT & BERTGraph-s & GCN & GraphConv \\
        \midrule
        Snopes  & \textbf{0.71} & 0.66 & 0.66 & 0.61  & 0.22\\
        N24News  & \textbf{0.69} & 0.64 & 0.57 &  0.49 & 0.46   \\    
        Politifact & 0.3 & 0.28 & 0.27 & 0.31 & \textbf{0.32}\\
        \bottomrule
      \end{tabular}
    }
\end{table} 

\subsubsection{Comparison between BERT based and GCN based models}
All three datasets are trained in two Graph Neural Network models, GCN~\cite{kipf2016semi} and GraphConv~\cite{morris2019weisfeiler}. They are compared only based on the accuracy. The dataset is trained for a maximum of 50 epochs with various layers for GCNs. Only the best-performing model is used for evaluation. Table 3 clearly shows that the accuracy of the GCN model depends on the dataset. GCN has $61\%$ accuracy on Snopes while GraphConv only has 22\% accuracy. The proposed model either outperforms GCN-based models or is comparable in accuracy. Besides the Politifact dataset, BERT-based models outperform GCN-based models. The pre-trained model captures general information from the sentences by embedding. The embeddings contain grammatical and structural information, contributing to their accuracy.

\subsubsection{Discussion}
In this project, we explore a crucial question that was evolving alongside the LLMs:
Can structured KBs enhance the performance of LLM’s downstream tasks, given that LLM’s outputs are unstructured?
Our experiments, as described in Subsections~\ref{exp:baseline} and Subsections~\ref{exp:training}, have yielded valuable insights.
We have observed that structural information plays a pivotal role during the fine-tuning process.
It can avoid the model overfitting on specific tasks.
The LLMs utilizing the attention mechanism, denoted as full attention, maybe overkill and excessive, leading to an overemphasis on input noise, which hurts the overall quality of the representations.
Structure information constrains the LLMs to fine-tune according to specific passes, encouraging the model to ignore the input noise.
This discovery holds the potential for LLMs to be more intelligent.

\section{Conclusion}
\label{sec:conclusion}
In this paper, we introduce the BertGraph frameworks and evaluate its effectiveness in downstream news-related classification tasks.
The BERTGraph is an extension of BERT, taking both original text and structured information extracted by NewsIE as its inputs.
Our evaluation demonstrates that, the BERTGraph outperforms its BERT Base counterpart in terms of almost all performance metrics for three news datasets available to the public, achieving up to $5\%$ of accuracy increase on the Snopes dataset.
The evaluation results underscore the importance of incorporating structural information for LLMs.
Furthermore, we observe that as the training dataset size grows, Bert suffers the over-fitting problem, which is not observed in BERTGraph.
Consequently, our model presents a more robust fine-tuning structure overall.
Additionally, we highlight the value of LLMs as rich sources of knowledge.
Our lightweight BERTGraph can easily utilize the implicit knowledge of LLMs.
In the future, we will evaluate BertGraph on larger datasets for a broader range of tasks, further exploring its potentials.


\bibliographystyle{ACM-Reference-Format}
\bibliography{main}


\end{document}